# BYPASSING CAPTCHA BY MACHINE—A PROOF FOR PASSING THE TURING TEST

***Ahmad B. A. Hassanat, PhD***
IT Department, Mu'tah University, Jordan

**Abstract**

For the last ten years, CAPTCHAs have been widely used by websites to prevent their data being automatically updated by machines. By supposedly allowing only humans to do so, CAPTCHAs take advantage of the reverse Turing test (TT), knowing that humans are more intelligent than machines. Generally, CAPTCHAs have defeated machines, but things are changing rapidly as technology improves. Hence, advanced research into optical character recognition (OCR) is overtaking attempts to strengthen CAPTCHAs against machine-based attacks. This paper investigates the immunity of CAPTCHA, which was built on the failure of the TT. We show that some CAPTCHAs are easily broken using a simple OCR machine built for the purpose of this study. By reviewing other techniques, we show that even more difficult CAPTCHAs can be broken using advanced OCR machines. Current advances in OCR should enable machines to pass the TT in the image recognition domain, which is exactly where machines are seeking to overcome CAPTCHAs. We enhance traditional CAPTCHAs by employing not only characters, but also natural language and multiple objects within the same CAPTCHA. The proposed CAPTCHAs might be able to hold out against machines, at least until the advent of a machine that passes the TT completely.

**Keywords:** Artificial intelligence, CAPTCHA, human interaction proofs, Turing test, optical character recognition

## Introduction

The Turing test (TT) scenario is based on an interrogator in one room chatting remotely with a machine and a human over some sort of network. The interrogator asks several questions to both human and machine and inspects their answers. If the interrogator can accurately differentiate between human and machine by virtue of the human's answers being more accurate, then the investigated system is not intelligent, and vice versa (Turing, 1950).

At the time of writing this paper, no artificial intelligence (AI) system has yet passed the TT using language as an indicator of intelligence. Language is perhaps the best indicator of intelligence, but it is not the only one—pattern recognition, for example, is another good indicator. Indeed, the questions in many IQ tests rely heavily on pattern recognition to measure human intelligence. Playing chess at a master's level is also an indication of intelligence.

In this context, we argue that any AI system that performs similarly to or better than a human is an intelligent system. This means that most AI systems designed to deal with problems that need extensive searching and many computations mostly beat human capabilities, because machines perform calculations much faster than humans. This applies to a large number of computer games, including but not limited to chess, n-puzzle, cube puzzle, n-queens, and crossword puzzles. Therefore, we believe AI has become much more intelligent than humans when applied in a specific domain. This means that the TT is satisfied, and is a proof for some AI solutions and intelligent models. If person A were to play chess online with person B on the other side of the world, and B used special chess software to determine the best moves to use against A, then A would never know that he was not playing against a human, because machines now play chess better than most humans.

Our claim is supported by Alan Turing himself. When he described machines, he stated that they *"are ugly, each is designed for a very limited purpose, when required for a minutely different purpose they are useless, the variety of behavior of any one of them is very small…"* (Turing, 1950).This means that machines are created for one purpose; if used for a different purpose, they may become useless. Each system is designed for a specific domain, and is good enough to replace humans in that domain only.

Based on the above, we may restate the TT as follows:

*For any specific well-defined task that requires human intelligence, if a machine's performance is equal to or greater than that of an averagely intelligent human, then the machine is intelligent.*

One application based on the theory that no computer can bypass the TT is Human Interaction Proofs (HIPs), commonly known as CAPTCHAs.

## Captcha

CAPTCHA is an acronym for "Completely Automated Public Turing Test to Tell Computers and Humans Apart."These were first proposed in 2000 (Von Ahn L., Blum, Hopper, & Langford, 2003) to describe a test that can differentiate humans from computers. The test must be easily read by humans, easily generated and evaluated, but not easily recognized by computers.

The main purpose of CAPTCHAs is to distinguish between humans and machines in security applications, such as preventing automatic comments and registration on different websites, protecting email addresses from web scraping, and protecting online polls from being skewed by certain views.

There are different types of CAPTCHAs, such as Gimpy and EZ-Gimpy, which are based on optical character recognition (OCR) (Von Ahn L. , Blum, Hopper, & Langford, 2003). Baffle text does not use English words, and combines images with circles, squares, ellipses, different fonts, and variable lengths and widths (Chew & Baird, 2003). Bongo is based on shape recognition (Von Ahn, Blum, & Langford, 2004), while Pessimal Print generates an image of random words with different fonts and image degradations (Coates, Henry, & Fateman, 2001). Another type of CAPTCHA uses audio with some added noise and distortion

The major challenge in designing CAPTCHAs is to make them easy enough for humans but difficult for computers. As AI advances, this gap becomes increasingly narrow (Azad & Jain, 2013). Thus, if an AI system could bypass a CAPTCHA, the website employing this system would be unable to differentiate between a human and a computer. In other words, the AI system is an intelligent system.

The main objective of this paper is to determine whether current OCR systems are intelligent in terms of the TT. A further objective is to prove that TT is a good intelligence test, regardless of its many criticisms (Saygin, Cicekli, & Akman, 2000) (Hayes & Ford, 1995). In addition, we propose a new generation of CAPTCHAs that cannot be broken by a machine that fails the TT.

For this purpose, we first review some work done to bypass CAPTCHAs. We then design and implement a simple OCR system that attempts to recognize randomly generated CAPTCHAs, and compare the results with answers obtained from a human responding from another room.

**Related work**

Breaking CAPTCHAs is not a new research theme. Mori and Malik have successfully broken EZ-Gimpy (with 92% success) and Gimpy (with 33% success) using methods based on shape context matching (Mori & Malik, 2003).

Moy et al. (Moy, Jones, Harkless, & Potter, 2004) developed a correlation algorithm that recognizes the word in an EZ-Gimpy challenge image with 99% accuracy, and a direct distortion estimation algorithm that recognizes the four letters in a Gimpy-r challenge image with 78% accuracy.

Chellapilla et al. (Chellapilla, Larson, Simard, & Czerwinsk, 2005) (Chellapilla & Simard, Using Machine Learning to Break Visual Human

Interaction Proofs (HIPs), 2004) compared human and computer single character recognition abilities by employing human users and computer experiments. These experiments assumed that the segmentation and approximate character locations were known. Their results show that computers are as good as or better than humans at single character recognition under distortion and clutter. In some of the reported experiments, computer and human performances were nearly indistinguishable.

A recent study by Serrao et al. (Serrao, Salunke, & Mathur, 2013) mentioned several strategies for bypassing CAPTCHAs. Some of these were developed for commercial purposes, and the reported results varied from 28% to 100% accuracy, depending on the software and the difficulty of the tested CAPTCHAs. Table 1 shows the success rates of some commercial CAPTCHA decoders reported in the study.

The deficiency of some traditional CAPTCHAs has led to the emergence of harder, and perhaps more intelligent, methods to protect websites from computer attacks. For instance, "Asirra" is a CAPTCHA that requires users to identify the cats in a set of 12 images of both cats and dogs (Elson, Douceur, & Howell, 2007).

In contrast to the above-mentioned works, (Coates, Henry, & Fateman, 2001) proposed a Pessimal Print CAPTCHA that uses degraded, low-quality images of machine-printed text synthesized pseudo-randomly over certain ranges of words. It was experimentally shown that the images are legible to human readers, but illegible to several of the best (at the time) OCR systems (see Figure 1, left). This work was motivated by a decade of research on the performance evaluation of OCR machines during the 1990s. Using three of what were then the best OCR machines, "*no machine guessed a single alphabetic letter, either correctly or incorrectly*" (Coates, Henry, & Fateman, 2001). Figure 1, right, shows how simple image processing (maximum, median filters followed by mean filter, then global thresholding) gives almost perfect segmentation, which normally leads to perfect recognition, because segmentation is the main problem when dealing with such computer vision tasks.

Table 1. Success rates of two commercial CAPTCHA decoders reported in (SERRAO, SALUNKE, & MATHUR, 2013)

We fully agree with the claims in (Coates, Henry, & Fateman, 2001) regarding the superiority of the Pessimal Print CAPTCHA, because the reported results were based on 1990s OCR technology. OCR and AI technologies are advancing rapidly. From the previous short review, we can see that CAPTCHA is no longer an effective means of distinguishing between machine and human.

| **Origin** | **Samples** | **Success rate** |
|---|---|---|
| Authimage | 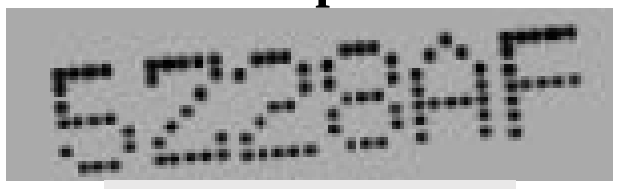 | 100% |
| linuxfr.org | 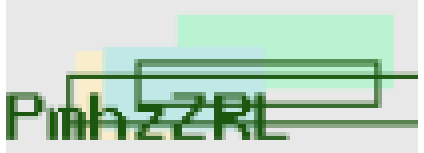 | 100% |
| Live Journal | 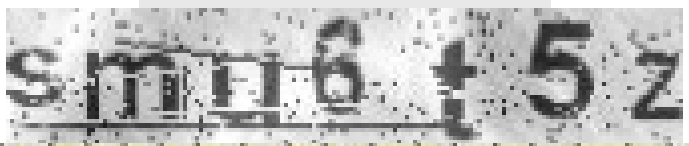 | 99% |
| Paypal | 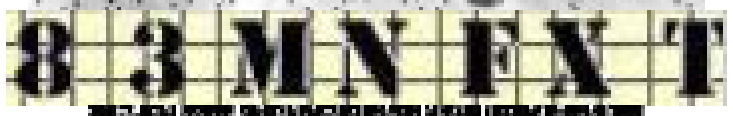 | 88% |
| vBulletin | 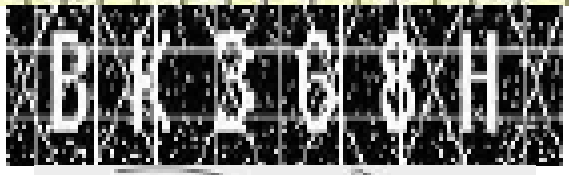 | 100% |
| Xanga | 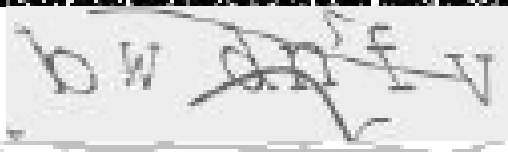 | 49% |
| Scode and derivatives | 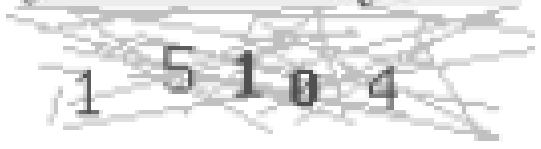 | 100% |
| Generic Bookmark/ RSS Directory | 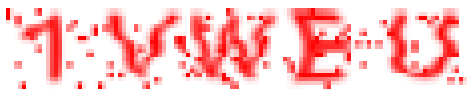 | 28% |
| Lifetime Blogs | 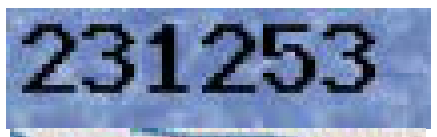 | 100% |
| Wordpress Blogs | 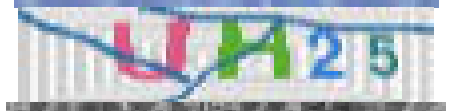 | 76% |
| Vivvo CMS | 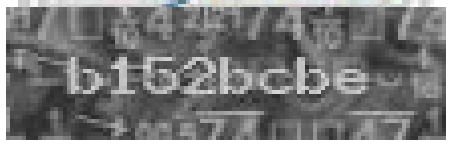 | 99% |

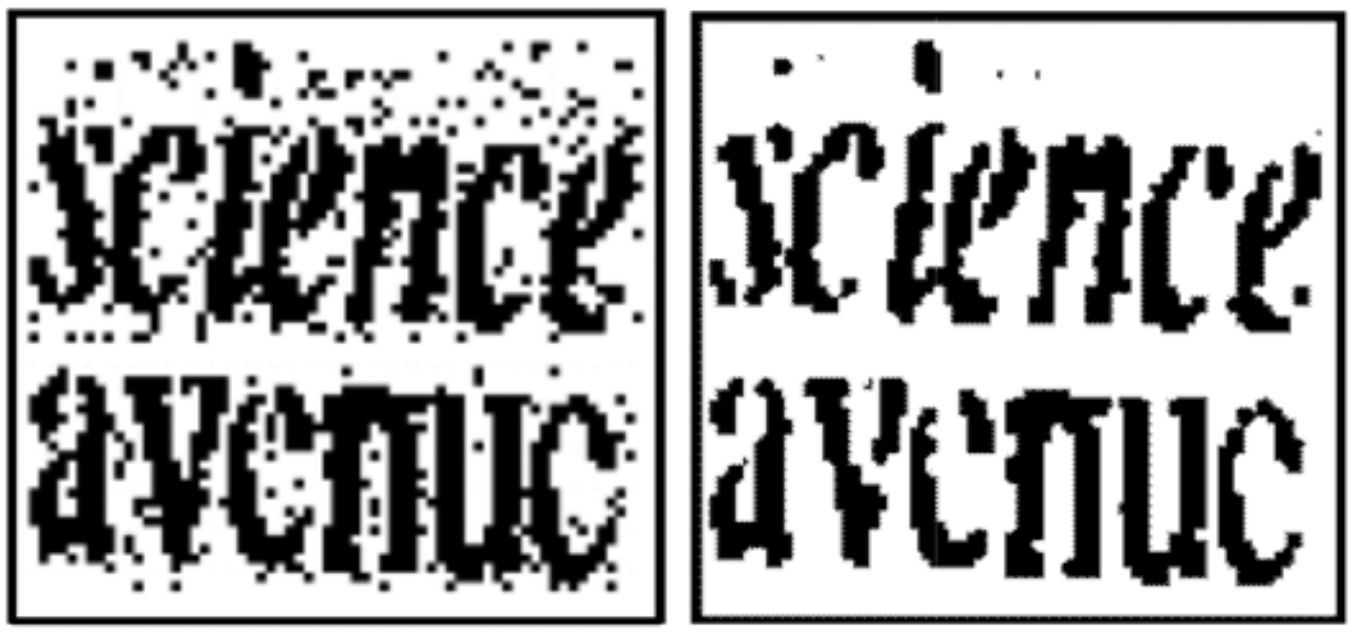

Figure 1. Left: Examples of synthetically generated images of printed words that are machine-illegible due to degradation (Coates, Henry, & Fateman, 2001). Right: Enhanced images.

## Simple OCR system used with the TT

In the proposed system, a CAPTCHA is created automatically using randomly chosen characters to create an image. This is then heavily degraded and sent over a network to a human and a client equipped with a simple OCR system. Both the human and machine read the CAPTCHA, and send the text back to the interrogator (see Figure 2).

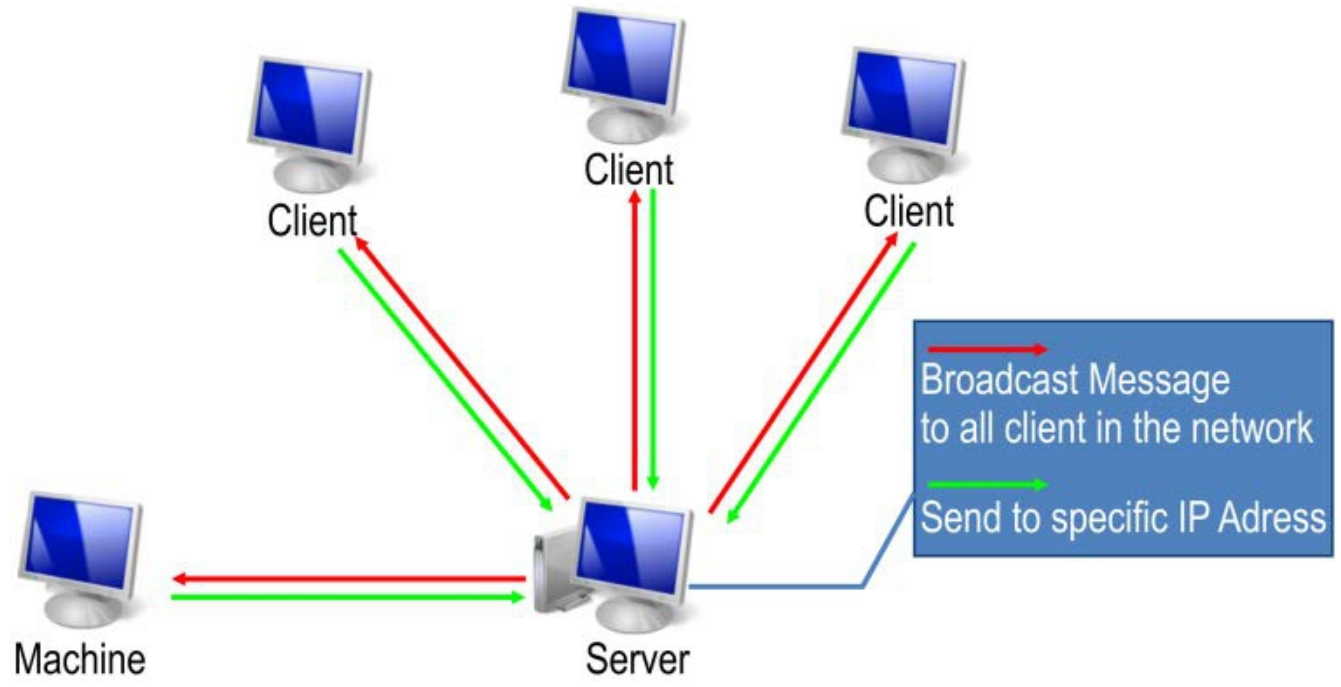

Figure 2. Simple OCR system used with the TT.

The judgment of the interrogator is based on accuracy—the most accurate answer is *assumed* to have come from the human. If the interrogator cannot distinguish between the two answers, i.e., the system's performance is comparable to the human's performance, then the system is intelligent in terms of this particular task.

The process of generating a random visual CAPTCHA is summarized in Figure 3.

| Process | Output |
|---|---|
| 1- Draw shadows for all characters | 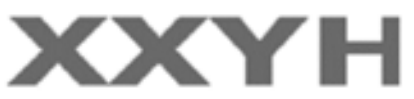 |
| 2- Draw random lines | 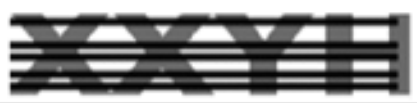 |
| 3- Draw the selected characters shifted left from the locations of the shadows | 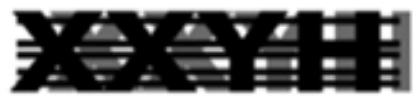 |
| 4- Add salt and pepper random noise to the image | 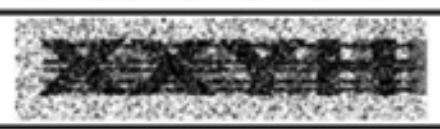 |

Figure 3. Process of creating a CAPTCHA out of the random characters XXYH.

The resulting CAPTCHA is sent to all clients, including the OCR machine, over a local area network. Both the humans and the OCR machine read/break the received CAPTCHA, and send the text (represented by the CAPTCHA) to the interrogator, who judges the answers based on their accuracy. The sequence of image processing and analysis performed by the OCR machine is summarized in Figure 4.

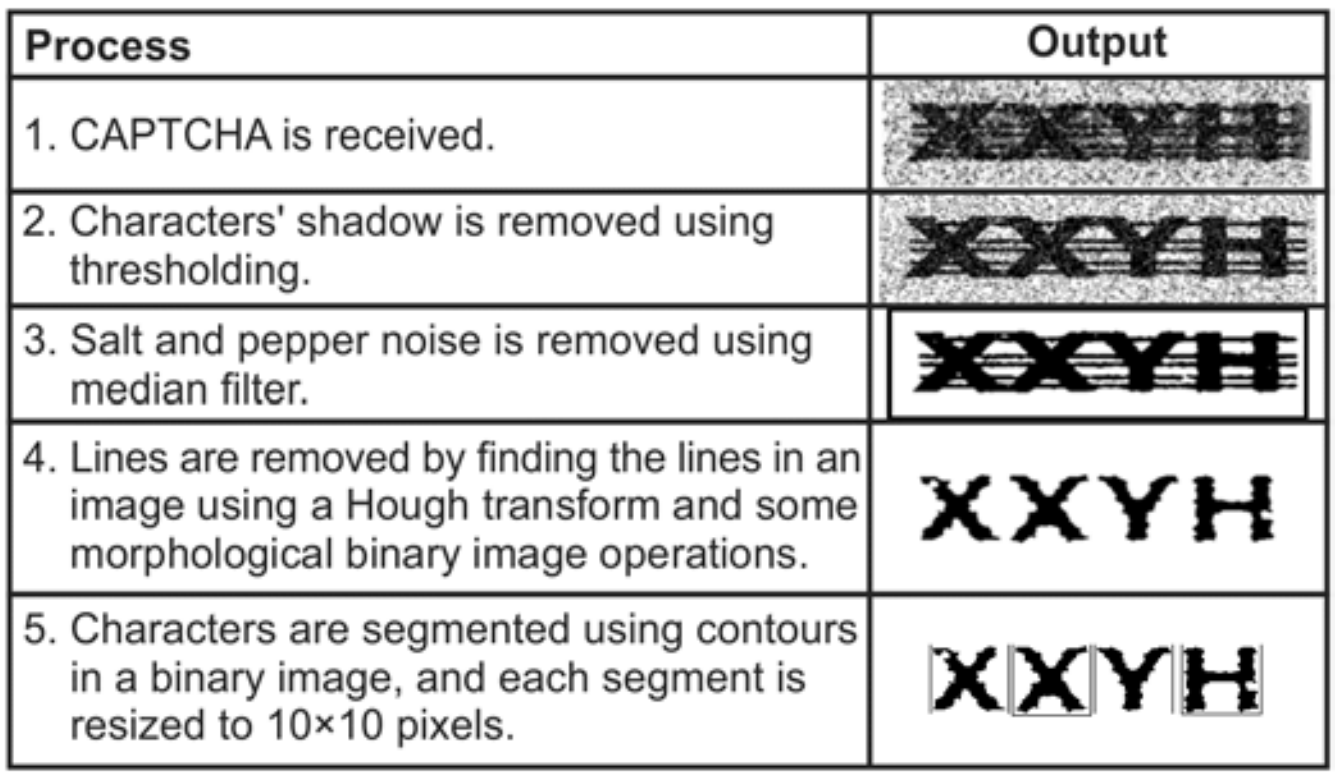

| Process | Output |
| --- | --- |
| 1. CAPTCHA is received. | |
| 2. Characters' shadow is removed using thresholding. | |
| 3. Salt and pepper noise is removed using median filter. | |
| 4. Lines are removed by finding the lines in an image using a Hough transform and some morphological binary image operations. | XXYH |
| 5. Characters are segmented using contours in a binary image, and each segment is resized to 10×10 pixels. | XXYH |

Figure 4. Sequence of image processing and analysis done by the OCR machine on the received CAPTCHA.

The OCR machine is trained on the English alphabet (upper and lower case in two different fonts) using an artificial neural network (ANN).Each segment is input to the ANN to find the best node index (that with the minimum error rate), which is associated with a specific character. All characters are concatenated and sent back (as a string) to the interrogator.

## Results and discussion

The above experiment was conducted 60 times. Each time, the same CAPTCHA was sent to a human and to the OCR machine. The human side was represented by six students from the IT department at Mutah University. Each student received 10 CAPTCHAs to read and the same 10 CAPTCHAs were sent to the machine. Table 2 reports the responses of the machine and the humans.

Table 2. Machine and human responses to the sent CAPTCHAs.

| Test # | CAPTCHA | Correct Answer | Machine Rate | Human Rate |
| --- | --- | --- | --- | --- |
| 1 | 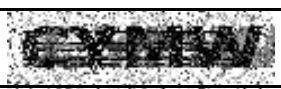 | CYMW | 4 | 4 |
| 2 | 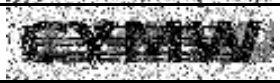 | CYMW | 4 | 4 |
| 3 | 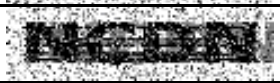 | NCDN | 4 | 4 |
| 4 | 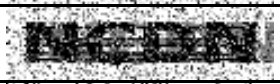 | NCDN | 4 | 4 |
| 5 | 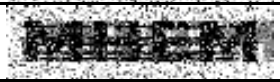 | MHEM | 4 | 4 |
| 6 | 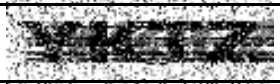 | YKTZ | 4 | 3 |
| 7 | 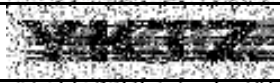 | YKTZ | 4 | 4 |
| 8 | 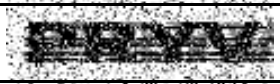 | PRYV | 4 | 4 |
| 9 | 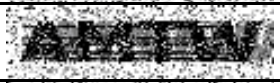 | AMTW | 4 | 3 |
| 10 | 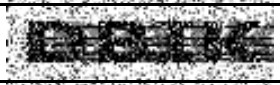 | DBIK | 3 | 4 |
| 11 | 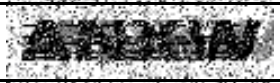 | ANXW | 4 | 4 |
| 12 | 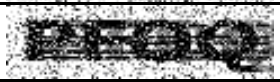 | PFOQ | 4 | 4 |
| 13 | 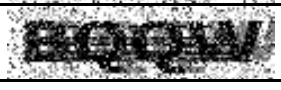 | BQQW | 4 | 4 |

| Test # | CAPTCHA | Correct Answer | Machine Rate | Human Rate |
|---|---|---|---|---|
| 14 | 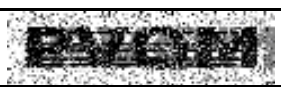 | PVOM | 4 | 4 |
| 15 | 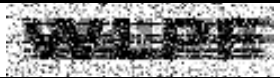 | WLPF | 4 | 3 |
| 16 | 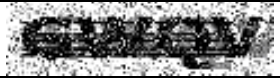 | GWQV | 3 | 4 |
| 17 | 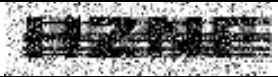 | HZNF | 4 | 4 |
| 18 | 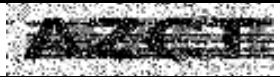 | AZCT | 4 | 4 |
| 19 | 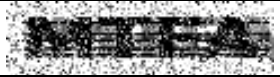 | MTFA | 4 | 4 |
| 20 | 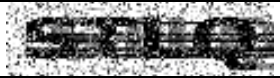 | STLQ | 4 | 4 |
| 21 | 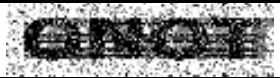 | ONOT | 4 | 4 |
| 22 | 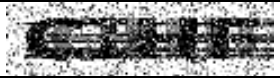 | CIHF | 3 | 4 |
| 23 | 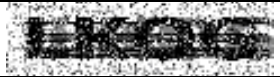 | LKOG | 4 | 4 |
| 24 | 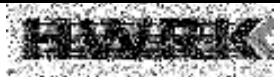 | HWRK | 3 | 4 |
| 25 | 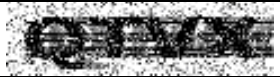 | QTVX | 2 | 4 |
| 26 | 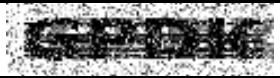 | GPDK | 3 | 4 |
| 27 | 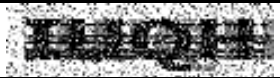 | IUQH | 4 | 3 |
| 28 | 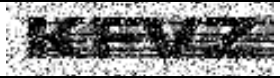 | KFVZ | 4 | 2 |
| 29 | 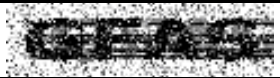 | GFAS | 3 | 3 |
| 30 | 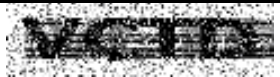 | VCTD | 3 | 3 |
| 31 | 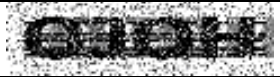 | OJOH | 4 | 3 |
| 32 | 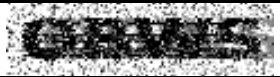 | OBWS | 4 | 3 |
| 33 | 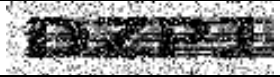 | DZPJ | 3 | 2 |
| 34 | 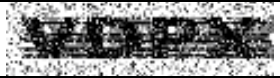 | VDPX | 3 | 3 |
| 35 | 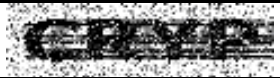 | CRYP | 3 | 4 |
| 36 | 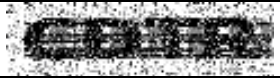 | CDTR | 3 | 2 |
| 37 | 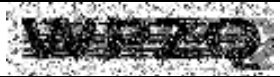 | WPZQ | 4 | 3 |
| 38 | 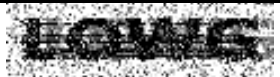 | LOWG | 4 | 3 |
| 39 | 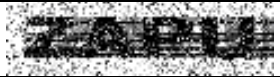 | ZAPU | 3 | 3 |
| 40 | 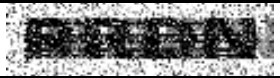 | RRDN | 1 | 3 |
| 41 | 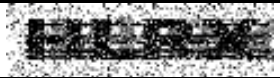 | HLRX | 3 | 0 |
| 42 | 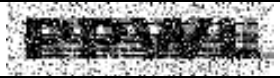 | PPWL | 4 | 1 |
| 43 | 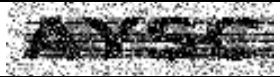 | AYSC | 4 | 3 |
| 44 | 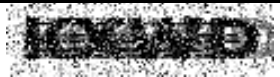 | KXWD | 3 | 4 |
| 45 | 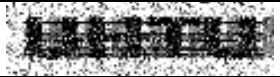 | UHTU | 4 | 0 |
| 46 | 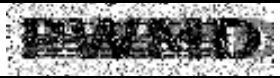 | PWMD | 4 | 4 |
| 47 | 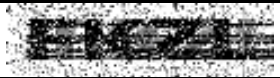 | FKZL | 4 | 3 |
| 48 | 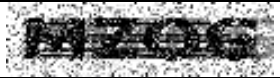 | MZOG | 4 | 3 |
| 49 | 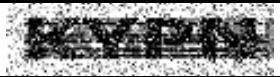 | KYPN | 4 | 4 |
| 50 | 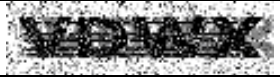 | VDWX | 3 | 4 |
| 51 | 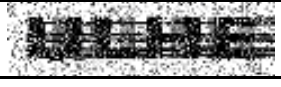 | ULHE | 4 | 3 |

| Test # | CAPTCHA | Correct Answer | Machine Rate | Human Rate |
|---|---|---|---|---|
| 52 | 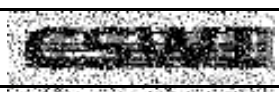 | CSWI | 4 | 3 |
| 53 | 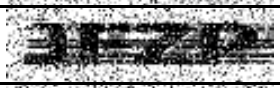 | JFZP | 3 | 4 |
| 54 | 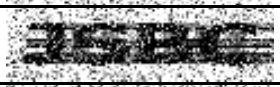 | JSBC | 4 | 3 |
| 55 | 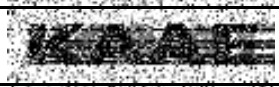 | KAAE | 3 | 4 |
| 56 | 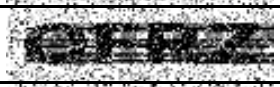 | OFRZ | 4 | 3 |
| 57 | 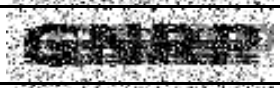 | GNRP | 2 | 3 |
| 58 | 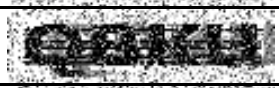 | QBKU | 4 | 3 |
| 59 | 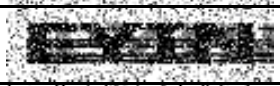 | EYIN | 3 | 4 |
| 60 | 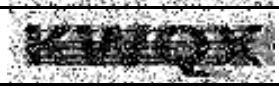 | KWQX | 4 | 4 |

As can be seen from Table 2, the interrogator could not distinguish between the humans and the machine based on the accuracy of their responses, as both are inaccurate (not intelligent) sometimes, and accurate (intelligent) at other times. The overall average single-character recognition rate was 89.58%for the OCR machine and 83.75% for the humans.

If the CAPTCHAs used in this experiment were used to protect a website, the machine would have bypassed it 65% of the time, and the humans would correctly access it 53.33% of the time.

Using such a simple OCR machine to bypass CAPTCHA places many websites in an unsafe situation, particularly those that use simple CAPTCHAs, such as the Jordanian National Database for Researchers (http://resn.hcst.gov.jo), which uses CAPTCHAs such as that depicted in Figure5.

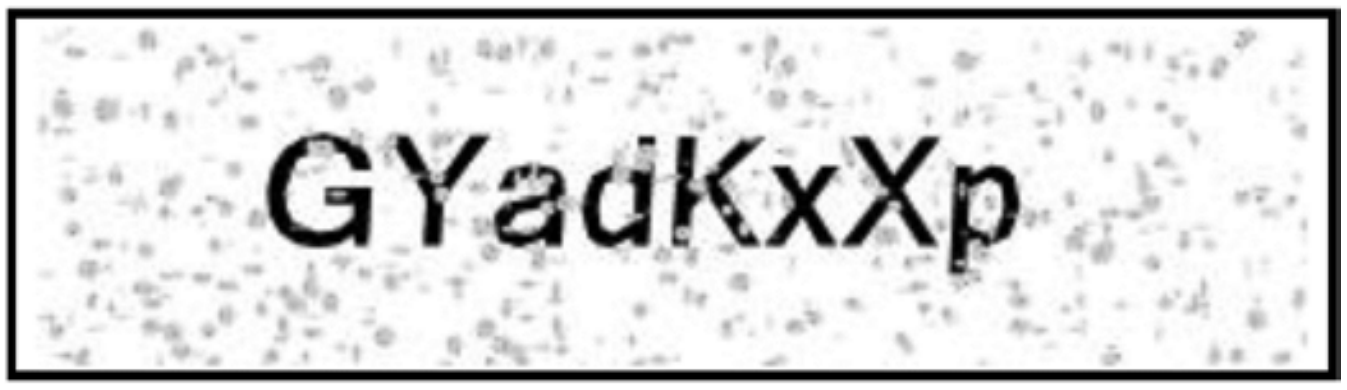

Figure 5. One example of a vulnerable CAPTCHA used to protect "http://resn.hcst.gov.jo" (Jordanian National Database for researchers)

We obtained 10 CAPTCHAs from http://resn.hcst.gov.jo, and conducted the same experiment using the same OCR machine and one student. In this experiment, the machine achieved a single-character recognition rate of 87.5%, and succeeded in recognizing the whole CAPTCHA in 70% of cases. The student was able to fully recognize all 10 CAPTCHAs. These results illustrate the weakness of some CAPTCHA systems that are based on weaknesses in OCR technology.

Instead of relying only on OCR as a basis for CAPTCHA, we propose an extended version based on natural language processing and general object recognition in addition to traditional OCR. This new CAPTCHA should be easily recognizable by humans but difficult for machines.

It is well known that machines cannot see all objects in an image, and there is no current system that can see and recognize large numbers of different objects. On the contrary, machines can only see within some specific narrow domain, such as OCR, face recognition, finger print recognition, and so on. Machines have an insufficient understanding of language, and natural language processing (NLP) is not mature enough to allow machines to understand everything. Nevertheless, there have been some successes in well-constrained problems.

Knowing the limited capabilities of machines in both general object recognition and the understanding of natural language, the extended CAPTCHA utilizes these deficiencies by producing an image containing different objects, randomly selected from a database containing thousands of objects, and different words arranged randomly around those objects. The system then asks the user to enter a specific character from a word identified by its location within the objects, or by its relation to some objects in the image. Figure 6 shows an example of the proposed CAPTCHA.

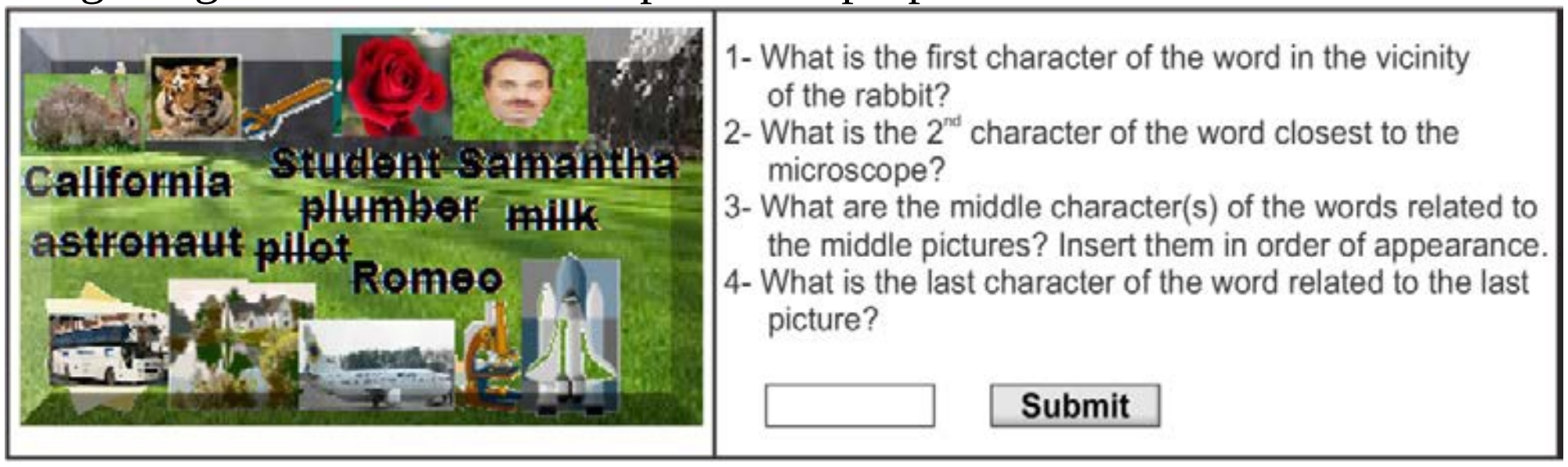

Figure 6. Example of the proposed CAPTCHA.

Figure 6 shows some typical questions that might be associated with the proposed CAPTCHA.

The answer should be "Comlt." The questions can be rephrased to be more complex and ambiguous in terms of NLP, but they should not reveal the contents of the image—this is important in holding off machine inference.

Answering these questions requires any AI system to have OCR, NLP (to understand the questions), and general object recognition and understanding abilities. Such a system is not currently available, and will not be available for at least 10 years, due to the complexity and ambiguity of such problems. This is particularly true when they are merged to form a more complex problem, such as in the proposed CAPTCHA.

If the performance of the employed OCR system was perfect (which is possible with current technology), the probability of guessing the required characters is given by:

$$P(all) = \left(\frac{1}{n}\right)^{l} \qquad (1)$$

where $n$ is the number of distinct characters in the used words, and $l$ is the number of questioned characters. Accordingly, the probability of guessing the CAPTCHA in Fig. 6 is $4.03861 \times 10^{-7}$.

If the system had sufficient NLP abilities, and could therefore understand the questions, the probability of guessing the previous CAPTCHA would be higher, but still very small—in this example, it would be $(1/8)^5 = 3.05176 \times 10^{-5}$,as there are eight words in Fig. 6. This probability can be reduced by including several characters for each question, using more words, more questions, and random words that include all ASCII codes.

## Conclusion

Degrading images by adding noise and breaking characters makes it more difficult, but not impossible, to solve CAPTCHAs with OCR programs. However, making something harder for a computer to recognize also makes it harder for humans.

The idea of CAPTCHA should no longer rely on the reverse TT, as we argue that TT is a good intelligence test, particularly for machines doing a specific task. In this instance, intelligent does not refer to absolute perception, cleverness, or innate intelligence. Rather, it means getting the job done accurately as if carried out by a human of average intelligence, i.e., AI.

If we simply consider getting the job done perfectly, is it still a matter of whether humans or machines are more intelligent? We end with Alan Turing's observation that "*We can only see a short distance ahead, but we can see plenty there that needs to be done*" (Turing, 1950).

At the time of writing, no AI system has passed the TT, because it requires language and image understanding. This is not currently possible, a situation that is unlikely to change in the near future. Therefore, we have suggested using such deficiencies to produce a new CAPTCHA based on OCR, NLP, and general image recognition.

Our future work will focus on the proposed CAPTCHA, with extensive investigations and further experiments.

## Acknowledgments

The author would like to thank Thaer Al-Aliat and Mahmoud Quran for applying the simple OCR machine and helping in conducting the experiments.